\PassOptionsToPackage{dvipsnames,table}{xcolor}
\documentclass[sigconf=true, nonacm=true, review=false, anonymous = false]{acmart}
\usepackage{subcaption}
\usepackage{amsmath}
\DeclareMathOperator{\mean}{mean}
\DeclareMathOperator{\std}{std}

\begin{document}

\newcommand{\review}[1]{\textit{\textcolor{purple}{ Review: #1}}}

\newcommand{\gina}[1]{\textit{\textcolor{purple}{ Gina: #1}}}
\newcommand{\todo}[1]{\textit{\textcolor{red}{#1}}}
\newcommand{\gapi}[1]{\textit{\textcolor{green}{Gasper: (#1)}}}
\newcommand{\tome}[1]{\textit{\textcolor{blue}{Tome: #1}}}
\newcommand{\carola}[1]{\textit{\textcolor{red}{Carola: #1}}}
\newcommand{\peter}[1]{\textit{\textcolor{brown}{Peter: #1}}}

\title{DynamoRep: Trajectory-Based Population Dynamics for Classification of Black-box Optimization Problems}

\author{Gjorgjina Cenikj}
\affiliation{
  \institution{Computer Systems Department\\ Jo\v{z}ef Stefan Institute\\ Jo\v{z}ef Stefan International Postgraduate School}
  \city{Ljubljana} 
  \country{Slovenia}
}
\email{gjorgjina.cenikj@ijs.si}

\author{Gašper Petelin}
\affiliation{
  \institution{Computer Systems Department\\ Jo\v{z}ef Stefan Institute\\ Jo\v{z}ef Stefan International Postgraduate School}
  \city{Ljubljana} 
  \country{Slovenia}
}
\email{gasper.petelin@ijs.si}

\author{Carola Doerr}
\orcid{0000-0002-4981-3227}
\affiliation{
  \institution{Sorbonne Universit\'e, CNRS, LIP6}
  \city{Paris}
  \country{France}}
\email{Carola.Doerr@lip6.fr}

\author{Peter Koro\v{s}ec}
\affiliation{
  \institution{Computer Systems Department\\ Jo\v{z}ef Stefan Institute}
  \city{Ljubljana} 
  \country{Slovenia}
}
\email{peter.korosec@ijs.si}

\author{Tome Eftimov}
\affiliation{
  \institution{Computer Systems Department\\ Jo\v{z}ef Stefan Institute}
  \city{Ljubljana} 
  \country{Slovenia}
}
\email{tome.eftimov@ijs.si}

\renewcommand{\shortauthors}{Cenikj, et al.}

\begin{abstract}
The application of machine learning (ML) models to the analysis of optimization algorithms requires the representation of optimization problems using numerical features. These features can be used as input for ML models that are trained to select or to configure a suitable algorithm for the problem at hand. Since in pure black-box optimization information about the problem instance can only be obtained through function evaluation, a common approach is to dedicate some function evaluations for feature extraction, e.g., using random sampling. This approach has two key downsides: (1) It reduces the budget left for the actual optimization phase, and (2) it neglects valuable information that could be obtained from a problem-solver interaction. 

In this paper, we propose a feature extraction method that describes the trajectories of optimization algorithms using simple descriptive statistics. We evaluate the generated features for the task of classifying problem classes from the Black Box Optimization Benchmarking (BBOB) suite. We demonstrate that the proposed DynamoRep features capture enough information to identify the problem class on which the optimization algorithm is running, achieving a mean classification accuracy of 95\% across all experiments.
\end{abstract}



\maketitle

\section{Introduction} 
Numerical ``\textit{feature-based}'' representations of black-box optimization instances can be used by Machine Learning (ML) models for \textit{automated per-instance algorithm selection} (determining the best algorithm for each problem~\cite{KerschkeHNT19}) and \textit{automated per-instance algorithm configuration} (finding the optimal hyper-parameters for each algorithm~\cite{BelkhirDSS17}). 
 Problem landscape features are also essential for ML-based analyses of the similarity of problem instances~\cite{urban_ela_not_invariant}, and the representativeness and redundancy of the problem landscape space in existing benchmark suites~\cite{selector,kate_space_filling_instances}.

 Testing the utility of problem landscape features for the task of classifying problems from the Black Box Optimization Benchmarking (BBOB) suite~\cite{bbob} is a common practice. Features characterizing problem instances using a subset of the Exploratory Landscape Analysis (ELA) features~\cite{bbob_classification_ela_feature_selection}, neural image recognition methods~\cite{bbob_problem_classification_neural_image_recognition}, and Topological Landscape Analysis (TLA) features~\cite{tla} have also been evaluated for the BBOB problem classification task. However, our work differs in the fact that the DynamoRep features proposed as part of this work, are extracted from the trajectory of an optimization algorithm, as opposed to samples of the entire problem landscape.

Features that capture properties of single-objective continuous optimization problems ~\cite{fla,tla,ela,dl_feature_free} can be roughly separated based on how they are obtained. Features such as Fitness Landscape Analysis (FLA)~\cite{fla} are most often defined by human experts based on the properties of the objective function which poses a problem for automatic algorithm selection/configuration. On the other hand, there are approaches that construct explicit features based on samples obtained from the objective function or construct features implicitly when solving some other task. The ELA features are calculated using a set of mathematical and statistical features applied on artificially sampled candidate solutions from the decision space of the problem instance. The drawback of ELA features is that they can be computationally expensive for high dimensional problems, and have been shown to be sensitive to the sample size and sampling method~\cite{ela_sensitive_to_sampling, urban_ela_not_invariant_sampling}, and are not invariant to transformations such as scaling and shifting of the optimization problem~\cite{urban_ela_not_invariant,urban_ela_not_invariant_sampling}.

All the previously described feature extraction methodologies construct optimization problem features based on a sample of candidate solutions from the entire optimization problem, i.e., they produce a ``static" representation of the problem which is not algorithm-specific. They capture properties of the entire problem landscape, regardless of which areas of the problem landscape have been visited by a specific algorithm, i.e., they may capture information about areas of the problem landscape that have never been visited by the algorithm. Furthermore, it has been shown that the parts of the problem landscape which an algorithm sees during the optimization process can differ substantially from the global fitness landscape of the problem being solved~\cite{adaptive_landcape_analysis}.

While such representations may be well-suited for problem or benchmark analysis, the tasks of automated algorithm selection and configuration could benefit from features that capture the interactions between the problem and the algorithm. Furthermore, using samples from the trajectory of the optimization algorithms means that there are no additional computational costs associated with evaluating the objective function of the optimization problem before running the algorithm.

In~\cite{adaptive_landcape_analysis}, Adaptive Landscape Analysis has been proposed, where the ELA features are computed using samples from the distribution that the algorithm (CMA-ES) uses to sample its solution candidates. Further, they have also been calculated using the candidate solutions visited by the algorithm during its run instead of using artificially sampled candidate solutions, not capturing the longitudinality of the solutions observed within the iterations of the algorithm execution. The approach has been evaluated for fixed-budget performance prediction~\cite{trajectory_ela_performance_regression} of the Covariance Matrix Adaptation Evolution Strategy (CMA-ES)~\cite{hansen1996adapting, hansen2016cma} algorithm, as well as \textit{per-run} algorithm selection with warm-starting~\cite{per_run_algorithm_selection_warmstarting}, where the features extracted from the trajectory of an initial optimization algorithm are used to determine whether to switch to a different algorithm.

\textbf{Our contribution:} In this paper, we propose a simplistic way of extracting features from the samples of the trajectories of an optimization algorithm using simple descriptive statistics. The proposed DynamoRep features capture the algorithm-problem instance interaction, not just global properties of the problem instance, and take into account the longitudinal nature of the algorithm trajectory. 
Furthermore, due to their simplicity, the proposed features have low computational overhead and require no additional sampling of the objective function. 
We evaluate the DynamoRep features for the task of identifying which of the 24 problem classes from the BBOB suite an algorithm is solving, using samples from the trajectories of the Differential Evolution (DE)~\cite{storn1997differential, pant2020differential}, Genetic Algorithm (GA)~\cite{katoch2021review}, Evolutionary Strategy (ES)~\cite{beyer2002evolution} and CMA-ES~\cite{hansen1996adapting, hansen2016cma} algorithms. We show that even when describing the search space of the algorithm with a few basic statistical measures, these measures carry enough information that can determine what problem the algorithm is solving. On average, we obtain a classification accuracy of around 95\% across all experimental settings. 
Additionally, we conduct an analysis of the performance of ELA features calculated from all candidate solutions visited by the algorithm run, for the problem classification task. We show that trajectory ELA features are not suited for this type of task when all candidate solutions are being considered, without taking into account the longitudinality of the trajectory data.

\textbf{Outline of the paper:} 
In Section~\ref{sec:methodology} we introduce the methodology to calculate the DynamoRep features from the search trajectories. The experimental design of our classification task is presented in Section~\ref{sec:experimental_design}. Results are presented in Section~\ref{sec:results}. 
Section~\ref{sec:conclusion} concludes our work and provides a discussion of future research directions.

\textbf{Reproducibility:} The full code is available at \url{https://github.com/gjorgjinac/DynamoRep_problem_classification.git}. This includes the code used to run the algorithms, extract the features, and perform the problem classification and analysis of the results. The algorithm trajectory data and the extracted features are available at Zenodo \url{https://doi.org/10.5281/zenodo.7598758}.

\section{Problem Representation Using Trajectory-based Population Dynamics}
\label{sec:methodology}

Given an optimization algorithm and an optimization problem instance, we aim to capture an algorithm-specific representation of the problem instance which takes into account the longitudinal nature of the algorithm trajectory. 

For the following description, we assume that we are dealing with a $d$-dimensional search space $S$ and a problem instance $f:S\rightarrow \mathbb{R}$. 
Denoting by $(x^i_1, x^i_2, ..., x^i_d, y^i)_{i=1,...\lambda}$ the $\lambda$ candidate solutions $\boldsymbol{x^i}=(x^i_1, x^i_2, ..., x^i_d)$ of the population and their function values $y^i=f(x^i)$ obtained in a single iteration, we use as a representation a concatenation of the following expressions\\
 $\left(\min\{ x^1_1,\ldots,x^\lambda_1\}, \ldots, \min\{ x^1_d,\ldots,x^\lambda_d\}, \min\{ y^1, \ldots, y^\lambda \} \right)$, \\ 
 $\left(\max\{ x^1_1,\ldots,x^\lambda_1\}, \ldots, \max\{ x^1_d,\ldots,x^\lambda_d\}, \max\{ y^1, \ldots, y^\lambda \} \right)$, \\ 
 $\left(\mean\{ x^1_1,\ldots,x^\lambda_1\}, \ldots, \mean\{ x^1_d,\ldots,x^\lambda_d\}, \mean\{ y^1, \ldots, y^\lambda \} \right)$, \\ $\left(\std\{ x^1_1,\ldots,x^\lambda_1\}, \ldots, \std\{ x^1_d,\ldots,x^\lambda_d\}, \std\{ y^1, \ldots, y^\lambda \} \right)$. \\ 
The representation concatenates the coordinate-wise minimum, maximum, mean, and standard deviation, respectively. This refers to the generation of a representation for a single iteration of the optimization algorithm. A representation for the entire trajectory is then generated by concatenating the representations for all of the iterations. Since we use four descriptive statistics, the representation for each iteration is of size $4(d+1)$. 
If the algorithm is run for $n$ iterations until the stopping criteria are met, the entire algorithm trajectory representation would then have a size of $4n(d + 1)$. One note is that since we are also extracting features from the first population created by the initialization procedure, we are also capturing properties of the whole search space. 

We chose these particular descriptive statistics to aggregate the dimensions of each population due to their low computational complexity and the fact that they can be obtained without additional function evaluations, by only using the candidate solutions evaluated by the algorithm during its execution.
With such aggregate statistics, we aim to capture the information on what kind of candidate solutions are generated by the algorithm (i.e. their range and diversity) in each iteration as well as the quality of the solutions. Moreover, such statistics are not sensitive to the sample size, can be calculated for any population, and are commonly used as aggregation operators in ML domains such as in computer vision~\cite{gholamalinezhad2020pooling}, and graph neural networks~\cite{corso2020principal}.

\section{Experimental Design}
\label{sec:experimental_design}
In this section, we introduce the key aspects of the experimental design. These include the \textit{problem portfolio} (the set of problem instances on which we run the algorithms and further use to perform classification), the \textit{algorithm portfolio} (the set of algorithms we run on the problem portfolio), and the \textit{machine learning model} (which we train for problem classification based on the proposed features). We further describe the experimental settings used to evaluate the models. Finally, we describe the baseline against which we are comparing the performance of the proposed features, i.e., the process of generating the trajectory ELA features. 

\textbf{Problem portfolio.} The Black-box Optimization Benchmarking (BBOB) suite~\cite{bbob} from the Comparing Continuous Optimizers (COCO) environment~\cite{coco} contains 24 single-objective optimization problem classes. Each problem class can further have multiple problem instances, which represent a transformation  of the original problem class. 
We use the first 999 instances of each problem class of dimension three, $d=3$, to demonstrate the applicability of the proposed representation for the problem classification task. 

\textbf{Algorithm portfolio:}
We use the \textit{pymoo}~\cite{pymoo} python library to run the Differential Evolution (DE), Genetic Algorithm (GA), Evolutionary Strategy (ES), and Covariance Matrix Adaptation Evolution Strategy (CMA-ES) algorithms on the problem instances from the BBOB benchmark. We run the algorithms in a fixed-budget scenario with a population size of $10d$. Since we are using three-dimensional problems, the population size is 30. We choose a relatively low budget of 30 iterations (900 function evaluations), since we would like to have features extracted from the initial iterations, in order not to exhaust a very high budget before the predictions are made. Note here that we are not at all interested in this study in the \textit{performance} of the solvers, nor how they compare to each other; our focus is on whether the DynamoRep features proposed in Section~\ref{sec:methodology} suffice for a reliable classification of the problem classes. The smaller the budget is, the better for our approach. 

All algorithms use Latin hypercube sampling to construct the initial population. The DE algorithm is configured to use random selection and binary crossover, with a crossover rate of 0.3. The rest of the configuration properties of the algorithms are set to the default values provided in the \textit{pymoo} library.

\textbf{ML classifier.} We train a Random Forest (RF) model to predict which of the 24 BBOB problem classes an algorithm is solving, given the proposed representation extracted from the trajectory of the algorithm run. We train a separate RF model for each of the optimization algorithms, meaning that the trajectories in the train set and in the test set are always obtained from the same optimization algorithm.
The RF model is executed using the default configuration parameters in the \textit{scikit-learn}~\cite{scikit-learn} python library. This means that the number of trees is set to 100, the gini impurity function is used to measure the quality of a split, and the tree nodes are expanded until all leaves are pure or until all leaves contain less than 2 samples. We do not perform parameter tuning to evaluate the impact of the proposed features only, on a fixed model configuration. The decision to use the random forest model was based on its strengths in handling tabular data~\cite{grinsztajn2022tree}.

The models are evaluated using stratified 10-fold cross-validation, where in each fold, 10\% of all problem instances from each problem class are removed from the training set and used for testing. For instance, in the first fold, problem instances 1-100 from each problem class are used for testing. 

\textbf{Experimental settings:}
We perform two experiments in order to test the generalization of an RF model trained for problem classification using features extracted from the trajectory of the optimization algorithm run. The experiments differ in how the data is split into train and test data sets. This has been performed to test the sensitivity of the features with regard to a different initial population used for the algorithm run (i.e., different random seeds).
\begin{itemize}
    \item The first experimental setting: the training data for the RF model always contains trajectories from the optimization algorithm run with a single random seed. The purpose of this experiment is to see if a model trained using data from a single run of the algorithm can generalize to other, unseen runs of the algorithm with different initial populations. Due to the stochastic nature of the algorithm, it could happen that the features are too dependent on the initial population and cannot be generalized if the algorithm is run with a different random seed and thus a different initial population.
    \item The second experimental setting: due to possible substantial differences in performance when the model is trained on trajectories from a single random seed, we also explore the generalization when a model is trained on trajectories from multiple runs of the algorithm with different random seeds. The idea behind this is to make the model less sensitive to the initial population. In this experimental setting, we use trajectories from four different runs of the optimization algorithm to train the RF model. 
\end{itemize}

\textbf{Analysis of the Performance of Trajectory-Based ELA features:} 
In our experiment, we have not performed a comparison with ELA features computed from each iteration due to several reasons. First, ELA features have been shown to be sensitive to the sample size, and most benchmark functions provide robust ELA values with a sample size in the range of $50d$, $100d$, $200d$~\cite{ryan_sample_size}. In our case, if we calculate them from each iteration, we only have 30 candidate solutions in each iteration, which is less than $50d$, $100d$, $200d$.  In the future, we are going to explore this with a larger population to guarantee robust calculation of the ELA features in this learning scenario.

However, we have conducted an analysis of the performance of trajectory-based ELA which have been previously proposed in the literature~\cite{adaptive_landcape_analysis, trajectory_ela_performance_regression}. The trajectory ELA features have been computed using samples from the entire algorithm trajectory. What this means is that all of the populations (i.e., samples of candidate solutions and their obtained solution values) from all of the iterations of each run of an optimization algorithm are concatenated and used as samples for computing ELA features. We need to point out here that this will not lead to a fair comparison, since the longitudinality is not taken into account. Additionally, the distribution of the candidate solutions in the problem landscape across the entire trajectory can be highly uneven, i.e., there are areas of the problem landscape that are over/underrepresented, which may be an issue for calculating some of the ELA features. For example, this happens when the algorithm is converging in the last iterations, and the samples are highly similar to each other, compared to the initial iterations. However, we add the results since this is a state-of-the-art approach for trajectory-calculated features. 

In particular, we compute ELA features from the following categories using the \textit{flacco} R library~\cite{flacco}: \textit{basic}, \textit{disp}, \textit{ela\_distr}, \textit{ela\_level}, \textit{ela\_meta}, \textit{ic}, \textit{nbc}, and \textit{pca}. In this way, we calculate a total of 93 features. Some of these features contain missing values, or have the same constant values for all trajectories and are not useful for training an RF model, so we discard these features. Some of the missing values happen for features since the matrix does not have a full rank so they cannot be computed.
We should note here that in this feature elimination process, all features from the categories \textit{ic}, \textit{nbc}, and \textit{ela\_distr} were removed, so only features from the categories \textit{basic}, \textit{disp}, \textit{ela\_level}, \textit{ela\_meta}, and \textit{pca} are used for training the problem classification model. After the feature elimination process, we have 46 features left for training the RF model.

\begin{figure*}[!t]
    \centering
    \includegraphics[width=0.8\linewidth]{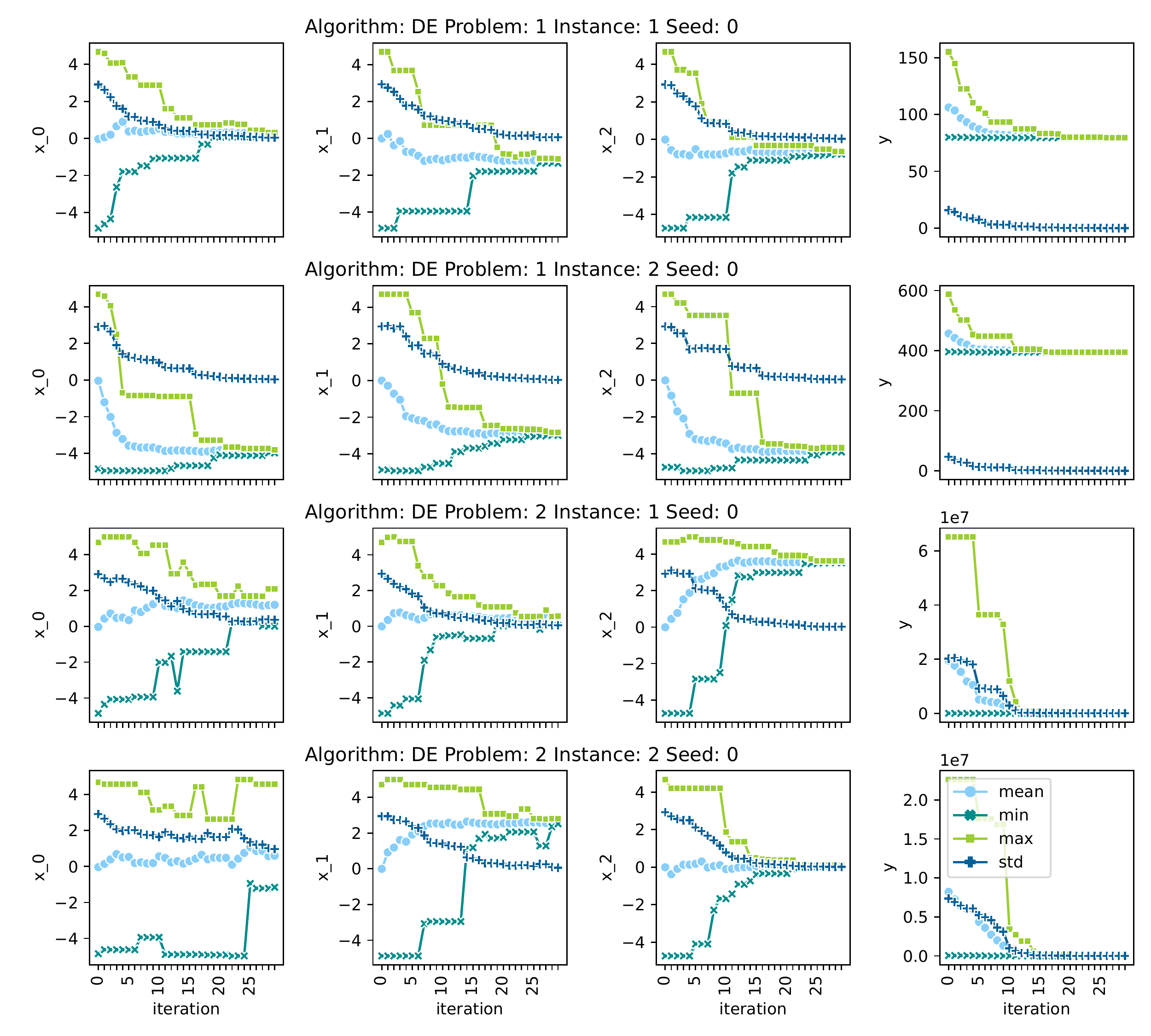}
    \caption{Line plots depicting the features generated from the trajectories of one run of the DE algorithm on the first two instances of the first two problem classes of dimension three from the BBOB benchmark suite. The first and second row show the first two instances of problem class 1, while the third and fourth row show the first two instances of problem class 2. The first column shows data for the first coordinate, the second column data for the second, etc., and the fourth column plots the evolution of the fitness values. The points at iteration 0 are the same for all three rows, since the seed is fixed to 0.  
    }    
    \label{fig:feature_plot_DE_pids_1-2_iids_1-2_seeds_0}
\end{figure*}

\section{Results and discussion}
\label{sec:results}
In this section, we present the results from the classification of the algorithm trajectories into one of the 24 BBOB three-dimensional problem classes using the proposed methodology. 
We first provide a visual analysis of the generated features. We then demonstrate in detail the applicability of the methodology for the classification of trajectories from the execution of the DE algorithm. To demonstrate its generalizability, we then shortly present the results from the classification of trajectories from three additional optimization algorithms: GA, CMA-ES and ES. We further analyze the performance of trajectory ELA features for the problem classification task, and conduct an error analysis of both approaches. 

\subsection{Exploratory Data Analysis}
Figure~\ref{fig:feature_plot_DE_pids_1-2_iids_1-2_seeds_0} presents plots of the generated features (i.e., mean, min, max, and standard deviation) from the trajectory of the DE algorithm. The first two rows contain the trajectories of the first two instances of problem class 1, while the third and fourth row show the first two instances of problem class 2.

Each column contains the features generated from one dimension of the problem ($x_0$, $x_1$, and $x_2$) and from the objective value ($y$). Each row contains features extracted from the trajectory of the DE algorithm run on a different problem instance. The y-axis shows the iteration from which each feature was extracted, while the x-axis shows the value of the feature. Each different colored line refers to a different feature type (light blue stands for mean, dark green for minimum, light green for maximum, dark blue for standard deviation). This shows how each of the features changes across the iterations of the algorithm.

At a first glance, the features look quite different for the different problem instances, which is most apparent in the ranges of the features generated from the objective function values (rightmost column, where the y-axis is labeled with "y"). This indicates that the problem classification task might not be trivial. However, while the mean, minimum and maximum features seem to differ across instances of the same problem class, the standard deviation feature seems to have a consistent behaviour across instances of the same problem class, indicating that this feature might be useful for differentiating between the problem classes.

\subsection{Classifying Trajectories from the Differential Evolution Algorithm}
In this subsection, we demonstrate the results from the application of the proposed methodology for the classification of the trajectories of the DE algorithm to the 24 three-dimensional problem classes in the BBOB benchmark. 

\begin{figure}
    \centering
    \includegraphics[width=\linewidth]{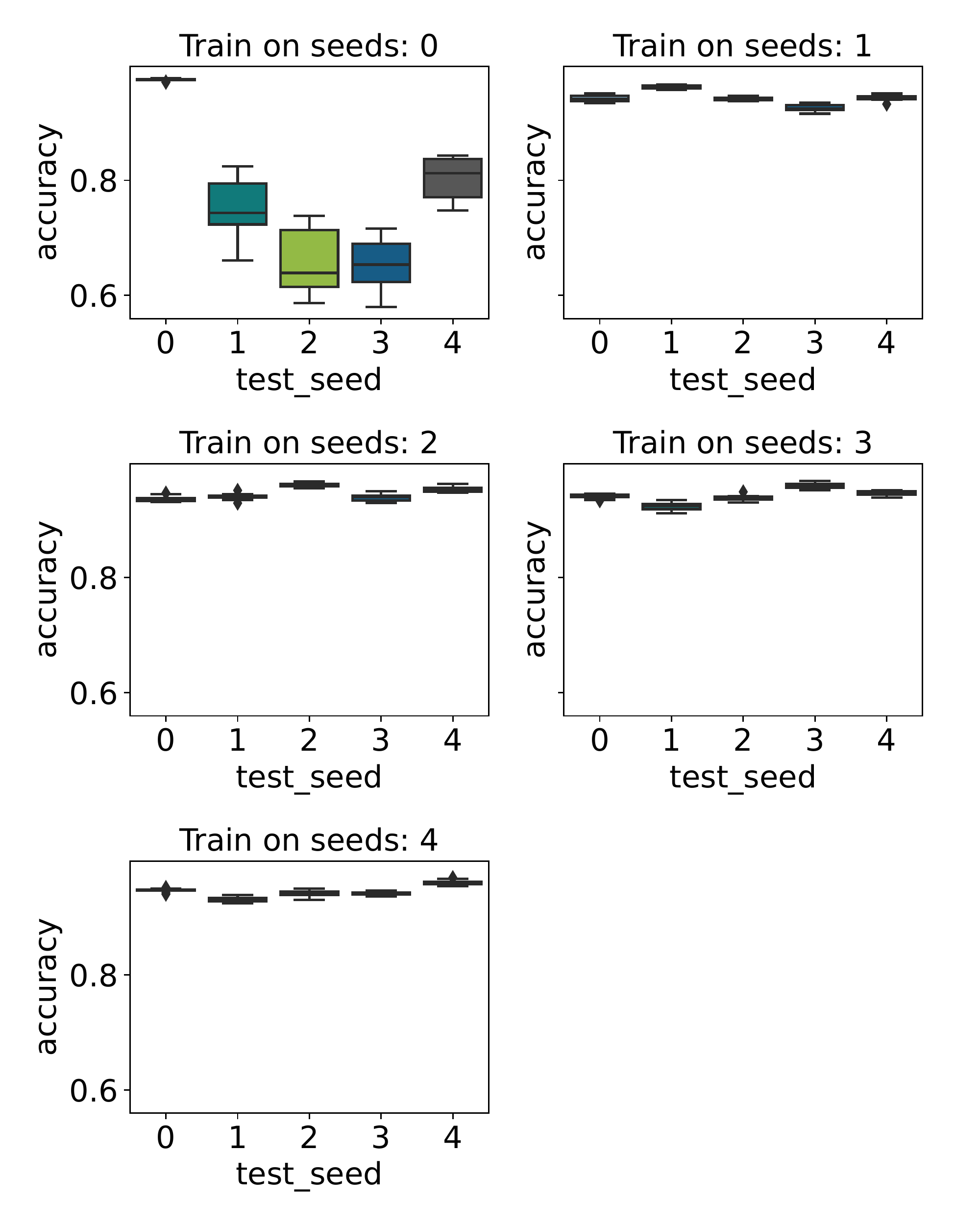}
    \caption{Boxplots representing the accuracy obtained in the 10-fold cross-validation of the RF models for the DE algorithm in the first experimental setting.
    In each subplot, we present the results when the training of the RF model is done using trajectories from the DE optimization algorithm executed with a single, different random seed. Each boxplot within a single subplot represents the result of the evaluation on a different random seed. The seed on which the model was trained is featured within the title of each subplot, while the seed on which the model was tested is featured on the x-axis of each subplot.}
    \label{fig:one_seed_results}
\end{figure}

Figure~\ref{fig:one_seed_results} presents the problem classification accuracy results obtained using 10-fold cross-validation, in the first experimental setting, i.e., when the training data contains trajectories from a single run of the algorithm executed using a single random seed and the testing data contains the trajectories from all five seeds. As we can see from Figure~\ref{fig:one_seed_results}, the models achieve the highest accuracy when tested on trajectories obtained when the algorithm was run with the same seed which was used for the trajectories in the training data. When tested on trajectories from different random seeds, the accuracy is somewhat lower. In some cases, as when the training is done on seeds 1, 2, 3, and 4, there is not a substantial difference in the accuracy obtained when trained on the same random seeds, and when trained on different random seeds. For instance, for train seed 1, the lowest median accuracy obtained when evaluated on a different random seed is 0.92. However, when the model is trained on seed 0, the lowest median accuracy on different random seeds is around 0.65. 

\begin{figure}
    \centering
    \includegraphics[width=\linewidth]{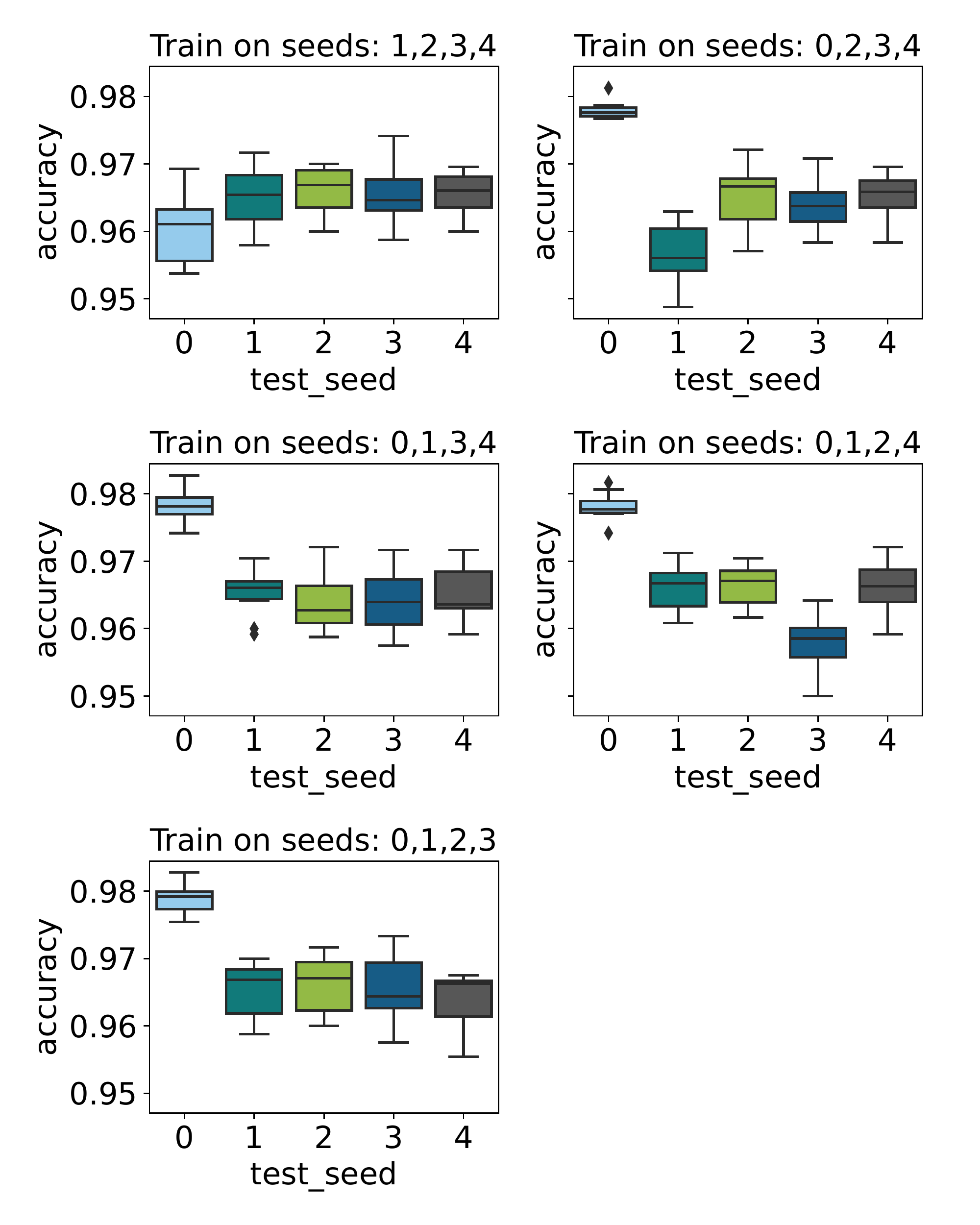}
    \caption{Boxplots representing the accuracy obtained in the 10-fold cross-validation of the RF models for the DE algorithm in the second experimental setting. In each subplot, we present the results when the training of the RF model is done using trajectories from the DE optimization algorithm executed with four different random seeds. Each boxplot within a single subplot represents the result of the evaluation on a different random seed. The seeds on which the model was trained are featured within the title of each subplot, while the seed on which the model was tested is featured on the x-axis of each subplot.}
    \label{fig:four_seed_results}
\end{figure}

 Figure~\ref{fig:four_seed_results} depicts the obtained accuracies in the second experimental setting, where the training data contains trajectories from four runs (four seed values) of the algorithm.
 The test data contains trajectories from all seeds, where 10\% of problem instances from each seed are used for testing. We show accuracies obtained on all five seeds, four where the model was trained on remaining 90\% of instances, and one seed that the model has never seen in training. In this case, we can see that the model generalizes better to different random seeds, with the lowest median accuracy of around 0.96. 

\subsection{Applicability to Other Algorithms}
In order to demonstrate the applicability of the proposed methodology on other optimization algorithms apart from DE, we repeat the previously presented experiments for the GA, ES, and CMA-ES algorithms. Table~\ref{tab:summary_accuracies} presents a summary of the mean and median accuracies obtained from the 10-fold cross-validation of the problem classification models trained on the proposed features extracted from the four optimization algorithms. In this case, we only present the generalization results, i.e., the trajectories from the runs of the optimization algorithm with a specific random seed are either used for training or for testing, but not for both. We conduct experiments where we use trajectories from a single run for training, and trajectories from four runs for testing (the first experimental setting), as well as experiments where we use trajectories from four runs for training, and trajectories from the remaining one run for testing (the second experimental setting). From Table~\ref{tab:summary_accuracies}, we can see that the results from all four algorithms are consistent with the initial experiments for the DE algorithm in that better accuracies are achieved with the second experimental setting. 
The lowest accuracy is achieved using trajectories from the DE algorithm, while the highest accuracy is achieved using trajectories from the ES algorithm. 

\begin{table}[]
    \centering
      \caption{Mean, median, and standard deviation of the accuracies obtained in the 10-fold cross-validation of the RF models trained and tested in the first and second experimental settings, using the proposed DynamoRep features.}
    \begin{tabular}{|p{40px}|p{50px}|p{32px}|p{32px}|p{32px}|}

 \hline
Algorithm &  Experimental Setting  &  Median  & Mean & Deviation\\ 
\hline
       DE &              1&         0.938651 &       0.895129 &                        0.096498 \\
       DE &              2 &         0.967083 &       0.967990 &                        0.006306 \\
       GA &              1 &         0.953333 &       0.953282 &                        0.010938 \\
       GA &              2 &         0.986250 &       0.986147 &                        0.002774 \\
    CMA-ES &              1 &         0.920000 &       0.920323 &                        0.009353 \\
    CMA-ES &              2 &         0.974167 &       0.974219 &                        0.004245 \\
       ES &              1 &         0.990833 &       0.989861 &                        0.004879 \\
       ES &              2 &         0.995417 &       0.995298 &                        0.001484 \\
       \hline

\end{tabular}
    \label{tab:summary_accuracies}
\end{table}

\begin{figure}
    \centering
    \includegraphics[width=\linewidth]{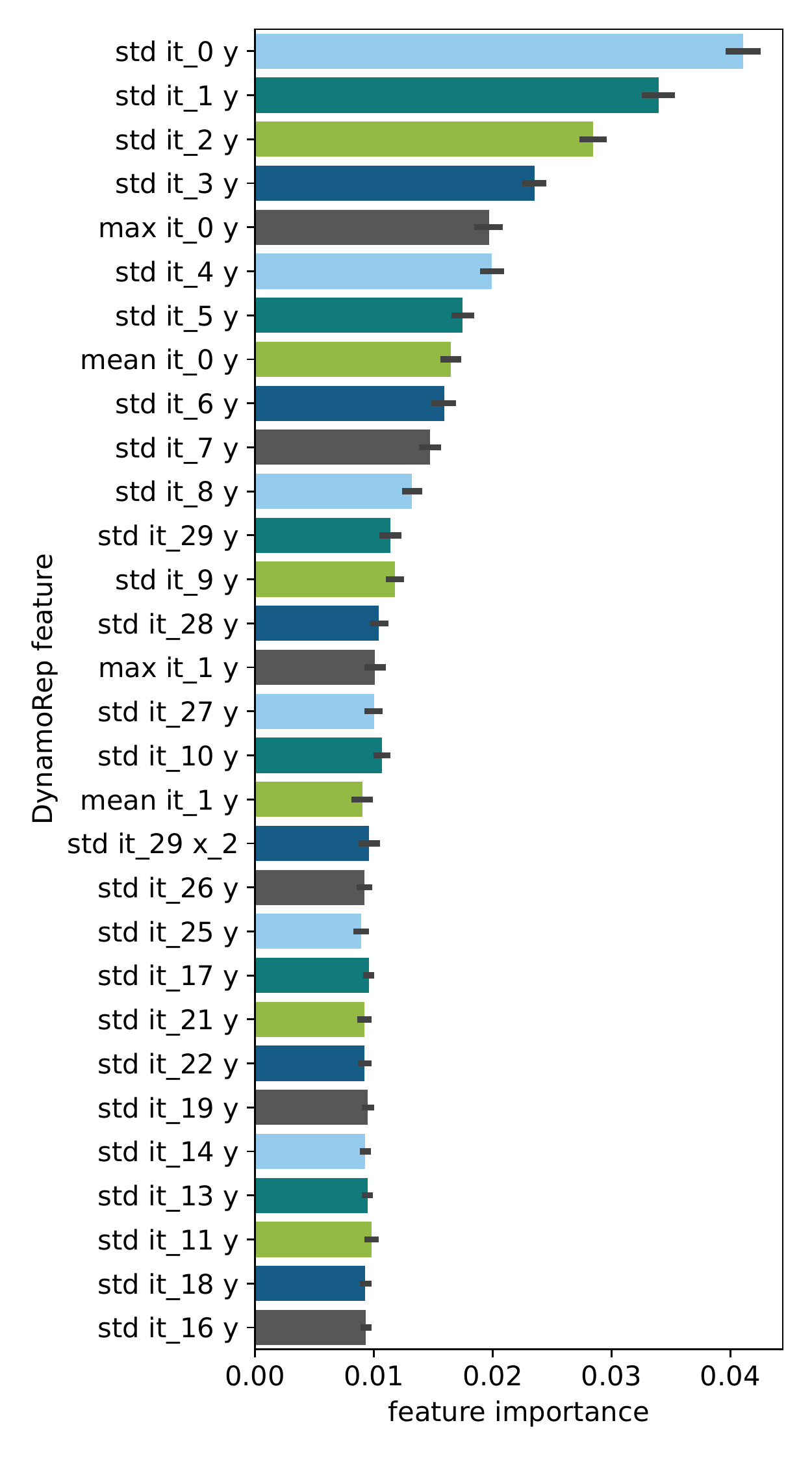}
    \caption{Top 30 features with highest median importance in the second experimental setting, aggregated across all folds, seeds and algorithms}
    \label{fig:top_30_features}
\end{figure}

To get a better insight into exactly the value of the proposed DynamoRep features, we analyze the importance values assigned to each of the features by the RF models. In Figure~\ref{fig:top_30_features} we plot the 30 features with the highest median feature importance. The feature importance is aggregated from every fold in the 10-fold cross-validation of the RF models performing classification of all of the trajectories of the DE, GA, CMA-ES and ES algorithms. Each barplot, therefore, represents an aggregation of 200 feature importance values (4 algorithms * 5 seeds * 10 folds). As can be seen from the figure, the top 18 features are all extracted from the values of the objective function. We can also note that 26 out of the top 30 features are obtained with the standard deviation statistic, and 25 out of those 26 features are extracted from the objective function values.

We can therefore conclude that the objective function value is more important that the candidate solutions, and the standard deviation is the most important statistic.

\subsection{Analysis of the Performance of Trajectory-Based ELA features}

Table~\ref{tab:ela_results} contains the mean and median accuracies obtained using the ELA features for all four optimization algorithms in both experimental settings. In this case, as in the previous section, we perform the evaluation on algorithm trajectories obtained with different random seeds than the ones used for training. This means that for the first experimental setting, the results in this section are obtained by evaluating on trajectories from four seeds, while in the second experimental setting, trajectories from one seed were used in the evaluation.
We can also observe that training on trajectories from four runs (the second experimental setting) provides substantially better results than training on trajectories from one run (i.e., the first experimental setting), with a mean improvement in the mean accuracies of 0.23. In this case, the classification done based on the trajectories of the ES algorithm provides the best results.

Compared to the corresponding results obtained with the DynamoRep features, which are depicted in Table~\ref{tab:summary_accuracies}, we can see that the results are considerably worse. In this case, the mean accuracy across both experimental settings and across all algorithms is approximately 0.49.

\begin{table}[]
    \centering
      \caption{Mean, median, and standard deviation of the accuracies obtained in the 10-fold cross-validation of the RF models trained and tested in the first and the second experimental setting, using the trajectory ELA features.}
    \begin{tabular}{|p{40px}|p{50px}|p{32px}|p{32px}|p{32px}|}
    \hline
Algorithm &  Experimental Setting  &  Median  & Mean & Deviation\\ 
\hline
       DE &              1 &         0.410625 &       0.410771 &                        0.063486 \\
       DE &              2 &         0.619167 &       0.617184 &                        0.037986 \\
       GA &              1 &         0.252500 &       0.274405 &                        0.070084 \\
       GA &              2 &         0.547083 &       0.556754 &                        0.037937 \\
    CMAES &              1 &         0.168969 &       0.177073 &                        0.051395 \\
    CMAES &              2 &         0.495417 &       0.491636 &                        0.028772 \\
       ES &              1 &         0.645208 &       0.593090 &                        0.131265 \\
       ES &              2 &         0.832706 &       0.832841 &                        0.009347 \\
\hline
\end{tabular}
\label{tab:ela_results}
\end{table}

This indicates that trajectory-based ELA features are not suitable for this kind of learning task, which is likely due to the fact that the ELA features are calculated on samples from the entire trajectory, not considering in which iteration the samples were generated. 

What we would also like to point out is that the computational cost associated with the calculation of the ELA features is much higher than the proposed DynamoRep features. Even if the ELA features did achieve a better predictive performance, the DynamoRep features already achieve mean classification accuracy of 95\% across all experiments, at a much lower computational cost.

\subsection{Error Analysis}
\begin{figure*}
    \centering
    \includegraphics[width=0.8\linewidth]{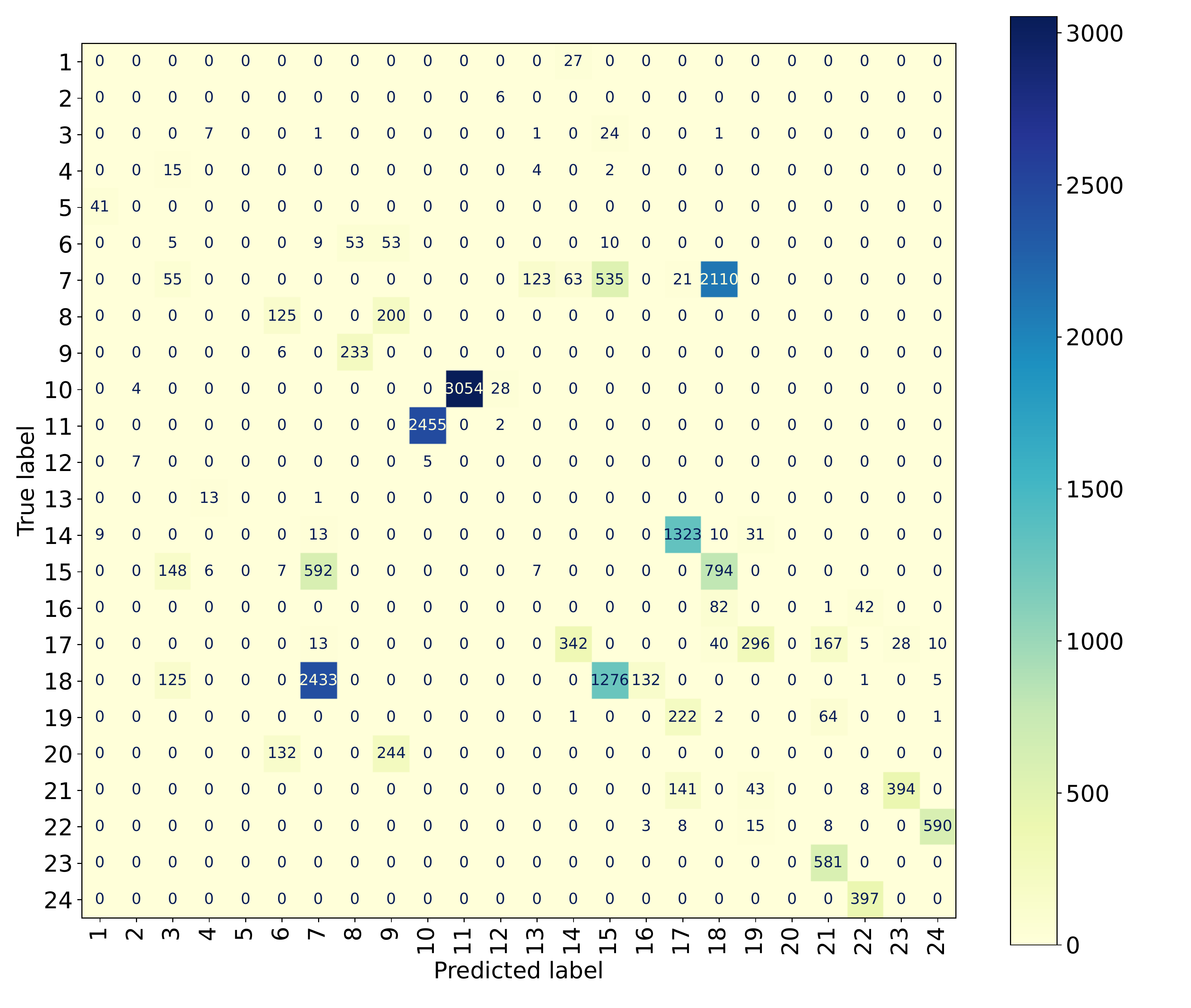}
    \caption{Confusion matrix obtained for the classification of DE trajectories using the proposed DynamoRep features}
    \label{fig:proposed_DE_confusion}
\end{figure*}

In this subsection, we analyze the errors made by the RF models in the problem classification. Figure~\ref{fig:proposed_DE_confusion} presents the confusion matrices obtained by the RF classifier using the trajectories generated by DE algorithm. 

We show results for the DE algorithm only, due to space limitations. The confusion matrices for the remaining algorithms, using both the DynamoRep and the ELA features, can be found in our github repository. The predictions are taken from the 10-fold cross-validation from the second experimental setting, when the testing data contains trajectories from all five runs of the algorithms. To highlight the model's mistakes in the figure, we are only presenting the incorrect predictions, so the numbers of the diagonal are all zeroes. 

As can be seen from the Figure~\ref{fig:proposed_DE_confusion}, when using the DynamoRep features, most problem instances are correctly classified. However, some problem classes are commonly confused, some examples being the pairs of problem classes 10 and 11, 7 and 18, 17 and 14, and 15 and 18.

\subsection{Runtime Analysis}
The experiments were executed on a system using the Ubuntu operating system, an Intel(R) Xeon(R) CPU E5-2680 v3 @ 2.50GHz, and 1 TB of RAM. On average, the training of the RF model takes 35.7 seconds in the first experimental setting, and 156.8 seconds in the second experimental setting (average execution time from 10 runs of the RF model training). The calculation of the DynamoRep features for one algorithm trajectory takes 0.02 seconds on average. The calculation of the features for a single run of one algorithm on all problem instances (999 instances of 24 problems, 23976 instances in total) takes 60 seconds on average. When scaling the approach to higher dimension, the calculation of the features themselves is not expected to be prohibitively high. However, before training the RF model, dimensionality reduction may be required since the dimensionality of the constructed features may be too large and may exceed the number of data instances used for training the model. Additionally, when the optimization algorithm converges in the last iterations, some of the constructed features are likely to not carry any meaningful information and may be omitted.

\section{Conclusions}
\label{sec:conclusion}
In this paper, we propose algorithm-specific optimization problem instance representations extracted from the populations generated by the optimization algorithm during its execution on the problem instance. Such representations capture longitudinal properties of the interactions between the algorithm and the optimization problem. We demonstrate that they can be used to distinguish between three-dimensional problem classes in the Black-Box Optimization Benchmarking suite, and they outperform the use of Exploratory Landscape Analysis features extracted from the entire algorithm trajectory. The proposed DynamoRep representations achieve a mean classification accuracy of 95\% across all experiments.
In this study, due to computational constraints, we limit ourselves to three-dimensional problem instances. Some directions for future analysis of the proposed methodology include testing the features for problem classification of problems of higher dimensions, as well as algorithm selection and algorithm configuration.

An important limitation of the proposed method is that the dimensionality of the representations is proportional to the dimension of the problem, meaning that for very high dimensions, the number of constructed features may be too large and may exceed the number of data instances used for training the model. In this case, dimensionality reduction techniques will be required to reduce the number of generated features. Initial experiments with problems of dimensions 5, 10 and 20 indicate that there is no considerable change in performance when increasing the dimension.

Additionally, in this paper, we have used a budget of 30 iterations, i.e. 900 function evaluations, but we did not investigate if smaller budgets would suffice. Computing the trade-off between classification accuracy and budget invested could be a worthwhile direction for future work.

Further analysis would be needed to assess if the models trained on trajectories of algorithms run with one population size would generalize to trajectories with different population sizes. Some of the proposed features are sensitive to the population size, and may cause the model not to generalize to other population sizes.

\section{Acknowledgements}
Funding in direct support of this work: Slovenian Research Agency: research  core  funding  No. P2-0098, young researcher grants No. PR-12393 to GC and No. PR-11263 to GP, projects No. N2-0239 to TE and No. J2-4460 to PK, and a bilateral project between Slovenia and France grant No. BI-FR/23-24-PROTEUS-001 (PR-12040). Our work is also supported by ANR-22-ERCS-0003-01 project VARIATION.

\bibliographystyle{ACM-Reference-Format}
\bibliography{references}

\end{document}